\documentclass[sigconf]{acmart}
\usepackage{booktabs} 
\usepackage{graphicx}
\usepackage{multirow}
\usepackage{placeins} 
\usepackage{subcaption}
\renewcommand\footnotetextcopyrightpermission[1]{}
\acmConference[Preprint]{}{}{}
\acmDOI{}
\acmISBN{}

\begin{document}
\title{Cracks in Embeddings: Robustness and Stability of JEPA-Based Chest X-Ray Representations}


\author{Zahid Mansur}
\affiliation{%
  \institution{Department of Computer Science and Engineering, IIT Jodhpur}
  \country{India}
}

\author{Kalpana Yadav}
\affiliation{%
  \institution{Department of Computer Science and Engineering, IIT Jodhpur}
  \country{India}
}

\author{Deepak Mishra}
\affiliation{%
  \institution{Department of Computer Science and Engineering, IIT Jodhpur}
  \country{India}
}

\renewcommand{\shortauthors}{Mansur, Yadav, and Mishra}

\newcommand{\dm}[1]{\textcolor{red}{[DM: #1]}}
\begin{abstract}
Self-supervised Joint Embedding Predictive Architecture (JEPA) models are increasingly used for medical imaging, however there is a limited understanding about robustness of these models toward real-world image changes. We test four JEPA-based chest X-ray encoders, trained on 78{,}468 NIH ChestX-ray14 images, under six tests: two black-box adversarial attacks (SPSA, Boundary Attack), three image corruptions (Gaussian noise, brightness shift, contrast shift), and a label-scarcity test, measuring embedding change and classification AUROC. These perturbations cause severe, model-dependent damage: encoder-level Cosine Similarity drops as low as 0.33, and AUROC falls by as much as 0.12 in the worst case (I-JEPA under the Boundary Test). We test two ways of training the small predictor on top of each frozen encoder to address this. Adversarial fine-tuning, trained on clean and perturbed images together, raises embedding consistency above 0.97 in nearly every case and improves AUROC, relative to the same perturbed baseline, for most models, with the largest gain of $+0.10$ on I-JEPA. A Vector-Quantized codebook method, trained only on clean images and evaluated on perturbed test images for the first time, instead degrades under distortion for all four models, most noticeably for I-JEPA ($-0.15$ AUROC on average), showing that its discrete bottleneck alone does not provide the robustness that comes from training on perturbed data directly.
\end{abstract}


\maketitle

\providecommand{\cellcolor}[2][]{}

\section{Introduction}

Deep learning has become the standard tool for analysis of medical images, however training these models usually needs large amounts of labeled data. In the medical domain, labeling is slow and expensive: it has to be done by clinical experts, and privacy rules limit how freely data can be shared, which makes fully supervised training hard to scale. Self-supervised learning (SSL) offers a way around for this problem. An SSL model first learns useful image features from a large pool of unlabeled images, and only a small labeled set is then needed to train the final task head. Among recent SSL approaches, the Joint Embedding Predictive Architecture (JEPA)~\cite{lecun2022path1,assran2023self} has drawn a lot of attention. JEPA relies on latent space predictions. It predicts the \emph{representation} of masked image regions from the visible part insted of predicting raw pixels. This pushes the model to capture high-level structure rather than low-level texture. This idea has now been carried into chest X-ray analysis, where several JEPA-style world models have been proposed, including CheXWorld~\cite{yue2025chexworld}, X-WIN~\cite{yang2026x}, and RadJEPA~\cite{khan2026radjepa}.

Strong accuracy on a clean test set, however, is not enough for a model that may be used in the clinic. Chest X-rays vary widely between scanners, hospitals, and acquisition settings, so the same anatomy can appear at different brightness, contrast, or noise levels. A deployed system may also be probed by an attacker who can query its outputs but not its internals. In all of these cases the image seen at test time differs from the clean training images, and a safety-critical model should degrade gracefully rather than fail silently. Whether today's JEPA chest X-ray encoders stay reliable under such changes, and what can be done when they do not, is still largely an open question.

Prior work has started to evaluate self-supervised medical encoders from different angles. A closely related effort is REMEDIS~\cite{azizi2022remedis}, which showed that self-supervised pretraining can improve the robustness and data efficiency of medical imaging models, but studied a single representation-learning recipe rather than comparing paradigms. Bundele et al.~\cite{bundele2024evaluating} went broader, running a systematic study of SSL in medical imaging across many datasets and methods (standardised through the MedMNIST benchmark~\cite{yang2023medmnist}) and covering label-efficient learning, cross-dataset generalisation, and out-of-distribution detection; they showed that choices such as the SSL method, the backbone, and the initialisation strategy strongly shape how well these models transfer, echoing earlier findings that transformer backbones benefit especially from pretraining in the medical domain~\cite{matsoukas2021replace}. These studies, however, do not measure how the learned representations behave when the input image itself is corrupted or attacked, and they offer no way to recover performance once it drops. Robustness to input perturbations has instead been studied mostly for natural images: through white-box adversarial attacks such as FGSM~\cite{goodfellow2014explaining} and the optimization attack of Carlini and Wagner~\cite{carlini2017towards}, through query-only black-box attacks such as SPSA~\cite{uesato2018adversarial} and the Boundary Attack~\cite{brendel2017decision}, and through common, non-adversarial corruptions~\cite{hendrycks2019benchmarking}; the underlying fragility of neural networks to small input changes has been known since Szegedy et al.~\cite{szegedy2013intriguing}. This line of work has not yet been connected to JEPA-based chest X-ray models.

Two gaps therefore remain. First, the input-perturbation robustness of JEPA chest X-ray encoders, under both realistic corruptions and black-box attacks, has not been measured, so robustness tends to be assumed rather than tested. Second, even where damage is found, as measured by a drop in classification AUROC value, there is no established way to train the downstream head to handle it. We keep each encoder frozen, attach a small predictor and classifier on top, measure how four existing JEPA chest X-ray encoders degrade under six test conditions, and then study two training methods that touch only this lightweight head: adversarial fine-tuning and a Vector-Quantised codebook~\cite{van2017neural}.

We make the following contributions:
\begin{itemize}
  \item We test four JEPA-based chest X-ray encoders under six conditions: two adversarial perturbations, three image corruptions, and a low-label test, using the full NIH ChestX-ray14 dataset, and measuring both embedding change and classification performance.
  
  \item We tested a structurally different training method based on Vector Quantisation (a discrete codebook) and compared it with the predictor that does not include a codebook.

  \item We design a fine-tuning method using the baseline predictor, which trains the predictor on clean and perturbed images together to improve the AUROC value.
\end{itemize}

\section{Related Work}

\textbf{JEPA models.} 
LeCun~\cite{lecun2022path1} proposed the JEPA idea: instead of rebuilding the raw pixels of a hidden part of an image, the model predicts that part's \emph{representation} in embedding space, using a small predictor network that looks at the visible part. This avoids wasting effort on small pixel-level details like exact texture or noise, and instead pushes the model to learn higher-level structure. I-JEPA~\cite{assran2023self} built this idea using a Vision Transformer: it hides a few large chunks of the image, then trains a predictor to guess what those chunks' representations should look like, based on the rest of the image, without needing any hand-designed data augmentations. CheXWorld~\cite{yue2025chexworld}, X-WIN~\cite{yang2026x}, and RadJEPA~\cite{khan2026radjepa} apply this same idea to chest X-rays, training their models to predict how a chest X-ray's representation changes across different views of the same anatomy, so the model learns the general structure of chest X-rays rather than just one narrow task. In these models, they are using a predictor that will predict the embeddings of missing patches. In our experiments, we use a predictor that follows the residual design, uses the same skip-connection idea as ResNet~\cite{he2016deep}, so it only has to learn a small adjustment on top of the frozen encoder's embedding. I-JEPA's own backbone is a Vision Transformer~\cite{dosovitskiy2020image}, the same type of architecture used by all four encoders in our study.

\textbf{Self-supervised representation learning.} Beyond JEPA, contrastive and distillation methods such as SimCLR~\cite{chen2020simple}, MoCo~\cite{he2020momentum}, and BYOL~\cite{grill2020bootstrap} learn representations without labels by comparing augmented views of the same image. Masked Autoencoders~\cite{he2021masked} instead rebuild masked pixel patches directly, a design JEPA avoids on purpose in favor of predicting in representation space.

\textbf{Evaluating self-supervised models.} Bundele et al.~\cite{bundele2024evaluating} tested self-supervised medical encoders for label-efficient learning and out-of-distribution detection, but not for robustness to corrupted or perturbed images, and gave no correction method.Our few-sample test (100\%, 10\%, and 1\% of training labels) follows the standard low-shot protocol used for self-supervised vision models~\cite{chen2020simple}.The encoders we study were pretrained on different datasets; for our own experiments, we use the NIH ChestX-ray14 dataset~\cite{wang2017chestx}.

\textbf{Black-box adversarial attacks.} We use two attacks that do not need access to model gradients, matching a realistic threat model where an attacker can only query a deployed model. SPSA finds a useful attack direction by random probing and reading the model's outputs~\cite{uesato2018adversarial}. The Boundary Attack starts from a large random change and slowly shrinks it while staying adversarial, using only the model's final decision~\cite{brendel2017decision}. Both differ from the earlier line of gradient-based, white-box attacks and defenses FGSM~\cite{goodfellow2014explaining}, the Carlini and Wagner attack~\cite{carlini2017towards}, and PGD-based adversarial training~\cite{madry2017towards}, the white-box counterpart to the fine-tuning approach we propose which all assume direct access to the model's internals rather than just its outputs. Beyond adversarial noise, Hendrycks and Dietterich~\cite{hendrycks2019benchmarking} showed that common, non-adversarial corruptions also hurt neural networks, which motivates our simple corruptions (noise, brightness, contrast) alongside the attacks. The general fragility of neural networks to small input changes was first pointed out by Szegedy et al.~\cite{szegedy2013intriguing}.


\section{Method}

\subsection{Overview}

Every encoder we test is frozen throughout the study. On top of it we attach two small trainable pieces: a predictor, which computes an adjusted output on top of it, and a classifier, which is used for disease prediction. Figure 1 summarises the study in two parts. Part (a) is training: all 78,468 training images pass through each of the four frozen encoders, and the resulting features train one predictor and classifier per encoder for 300 epochs on clean images only; this pair is the ``Original'' baseline used everywhere else in the paper. Part (b) is the perturbation-and-test loop used for every result in Section 5: held-out test images go through the five perturbation tests, through the same frozen encoders, and then through two separate predictor variants built on the Original predictor: a fine-tuned predictor/classifier (Section 3.5.1) and a codebook predictor/classifier (Section~\ref{sec:codebook}). Both variants are evaluated on the same distorted test images, so they can be compared fairly. In all cases the encoder itself is never trained or changed; only the small predictor and classifier are trained and tested.

\begin{figure*}[t]
  \centering
  \includegraphics[width=0.85\textwidth]{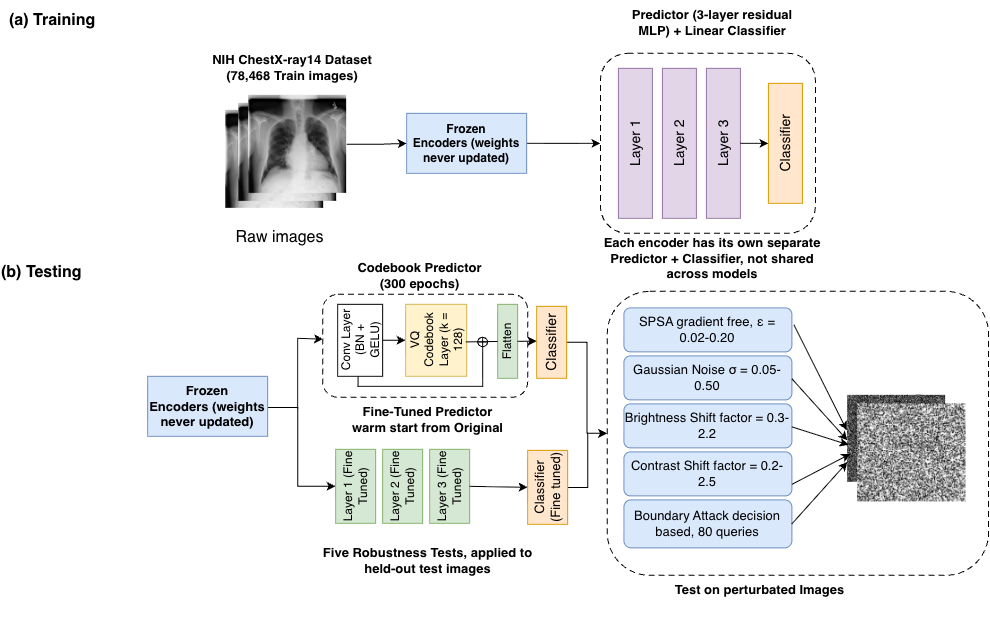}
  \caption{(a) Training: each frozen encoder trains its own predictor/classifier pair on clean images, not shared across models. (b) Test: the held-out set is perturbed, passed through the same frozen encoders, and evaluated by two independently trained predictors, codebook (clean images only) and fine-tuned (clean and perturbed images).}
  \Description{Two-part schematic. (a) Training: chest X-ray dataset, four frozen encoders with their embedding sizes and pretraining sources, a predictor and classifier, producing the clean-trained baseline. (b) Test: held-out test images, a box listing the five robustness tests and severity ranges, the same four frozen encoders, then two branches, a codebook predictor/classifier and a fine-tuned predictor/classifier, evaluated on distorted images.}

  \label{fig:pipeline}
\end{figure*}

\subsection{Predictor Architecture}
\label{sec:predictor}

The predictor is a small residual network:
\begin{equation}
  \label{eq:predictor}
  z_{p} = z + W_3 \,\sigma\!\big(W_2\, \sigma(W_1 z)\big)
\end{equation}

Where, $z$ is the raw embedding produced by the frozen encoder and never changes; $W_1, W_2, W_3$ are the three trainable weight matrices of the predictor (sized $D{\times}512$, $512{\times}512$, and $512{\times}D$, where $D$ is the encoder's embedding size); $\sigma$ is the GELU activation; $z_p$ is the predictor's output, passed on to the classifier. As Equation~\ref{eq:predictor} shows, $z_p$ is computed by adding the original embedding $z$ directly to the output of the three trainable layers a residual (skip) connection~\cite{he2016deep}. Because of this, the predictor only has to learn a small adjustment on top of $z$, rather than reconstructing the whole embedding from scratch.

The classifier is a single linear layer, $\hat{y} = W_c\, z_p + b_c$, where $W_c$ and $b_c$ are trainable weights and $\hat{y}$ is the predicted probability for each of the 14 chest X-ray disease labels.

The predictor and classifier are trained together, using only clean (unperturbed) images. After this training, both are frozen and used as the ``Original'' baseline for every robustness test in this paper.

\subsection{Measuring Robustness}

To test robustness, we take a clean image $x$, apply a perturbation function $A$ (an attack or corruption) at some severity level to get a perturbed version of it, and compare the results on the clean image versus the perturbed image, using two measures: how much the predictor's and classifier's predictions change.
\begin{equation}
  \label{eq:cossim}
  \mathrm{CosSim}(z_p,\, \tilde{z}_p) = \frac{z_p \cdot \tilde{z}_p}{\lVert z_p \rVert \, \lVert \tilde{z}_p \rVert}
\end{equation}

Where, $z_p$ is the predictor's output embedding for the clean image; $\tilde{z}_p$ is its output for the perturbed version of the same image; $\mathrm{CosSim}$ ranges from $-1$ to $1$, where $1$ means the two embeddings point in exactly the same direction (the perturbation caused no change) and lower values mean it pushed the embedding further from its original position. We also measure Macro AUROC of the classifier's predictions on both the clean and the perturbed embedding, using the true disease labels.

\subsection{The Six Robustness Tests}

We test five perturbations at four severity levels each (the Boundary Attack has one fixed setting), plus one label-scarcity test. Table~\ref{tab:attack-config} lists every severity level used.

\begin{itemize}
  \item \textbf{SPSA}: SPSA is a gradient-free attack~\cite{uesato2018adversarial}. It does not have access to the model's internal gradients, only its outputs. At each step, it tries a small random change to the image and checks whether that change pushed the model's output embedding further from the original. It uses this information to estimate which direction is most damaging. Repeating this process over many steps builds up a full, effective perturbation.

\item \textbf{Boundary Attack}: The Boundary Attack~\cite{brendel2017decision} begins with a large random change to the image, large enough to flip the model's prediction. It then repeatedly shrinks the change by a small step, checking each time whether the prediction is still flipped. If it is, it keeps the smaller change and continues; if not, it steps back. It stops once no further shrinking is possible without the prediction flipping back.

\item \textbf{Gaussian Noise}: This corruption adds random noise, drawn from a Gaussian (normal) distribution, to every pixel of the image. It simulates the kind of sensor noise that real imaging hardware can introduce.

\item \textbf{Brightness Shift}: This corruption multiplies every pixel value by a fixed factor, making the whole image uniformly darker or brighter. It simulates exposure differences between scanners.

\item \textbf{Contrast Shift}: This corruption stretches or shrinks pixel values around the image's average brightness. It imitates contrast differences between scanners. Together, these three corruptions follow the spirit of common-corruption robustness benchmarks~\cite{hendrycks2019benchmarking}, adapted here to chest X-ray conditions.
\item \textbf{Few-Sample Robustness Test}: This test does not perturb any image at all. Instead, it trains a fresh linear classifier, with no predictor involved, using only 100\%, 10\%, or 1\% of the training labels. It measures how good the encoder's raw embeddings are when very little labeled data is available.
\end{itemize}

\begin{table}[t]
  \caption{Severity levels used for each perturbation test.}
  \label{tab:attack-config}
  \small
  \begin{tabular}{ll}
    \toprule
    Test & Severity levels tested \\
    \midrule
    SPSA & $\epsilon = 0.02,\ 0.05,\ 0.10,\ 0.20$ \\
    Boundary Attack & fixed (80 queries) \\
    Gaussian Noise & $\sigma = 0.05,\ 0.15,\ 0.30,\ 0.50$ \\
    Brightness Shift & factor $= 0.3,\ 0.6,\ 1.6,\ 2.2$ \\
    Contrast Shift & factor $= 0.2,\ 0.4,\ 1.8,\ 2.5$ \\
    Few-sample & $100\%,\ 10\%,\ 1\%$ of sample images \\
    \bottomrule
  \end{tabular}
\end{table}

We use only black-box attacks (SPSA, Boundary), which do not need access to model gradients, since a realistic attacker on a deployed medical system usually has query access only, not internal access, unlike white-box attacks such as FGSM~\cite{goodfellow2014explaining}.

\subsection{Two Training Methods for the Predictor}
Since the encoder is frozen, the only way to check how the model reacts to a distorted image is by how the predictor and classifier on top of it are trained. 

\subsubsection{\textup{\textbf{Adversarial Fine-Tuning}}}

\label{sec:ft}

The goal of this method is simple: teach the predictor to give the same answer on a clean image and its perturbed version. We start from the clean image to train the predictor and classifier for the baseline (Section~\ref{sec:predictor}), but now, for this adversarial fine-tuning, we use a mix of clean and perturbed images instead of clean images only. For each image in the training set, we generate one perturbed copy for every perturbation type listed in Table~\ref{tab:attack-config}, always using that perturbation's strongest (most damaging) severity level.
This is similar in spirit to adversarial training~\cite{madry2017towards}, except we only train the small predictor head on top of a frozen encoder, not the whole network.

The training loss has three parts, one for each thing we want the model to do:

\begin{equation}
  \label{eq:ftloss}
  \mathcal{L} = \underbrace{\tfrac{1}{2}\mathrm{BCE}(\hat{y}, y) + \tfrac{1}{2}\mathrm{BCE}(\tilde{\hat{y}}, y)}_{\text{classify both correctly}} \; + \; \underbrace{\lambda\big(1 - \mathrm{CosSim}(z_p, \tilde{z}_p)\big)}_{\text{keep embeddings close}}
\end{equation}

Where, $\hat{y}$ and $\tilde{\hat{y}}$ are the classifier's predictions on the clean and perturbed image; $y$ is the true label; $\mathrm{BCE}$ is binary cross-entropy (the standard loss for multi-label classification); $\lambda$ is a weight (we use $\lambda{=}1$) controlling how strongly we enforce embedding similarity. The first two terms push the model to classify correctly whether or not the image was perturbed. The third term penalises the model whenever the perturbed embedding drifts away from the clean embedding, so the predictor learns to treat both versions of the same image as the same thing.

For efficiency, the SPSA and Boundary Attack caches used during fine-tuning are built on a large subset of the training split rather than the full set, since these two attacks require repeated forward passes per image and are far more expensive to generate than the simple corruption-based attacks. The simple corruptions (Gaussian noise, brightness, contrast) are instead computed over the entire training split, since a single extra forward pass per image keeps their cost low. This reduced budget is used only to build training data. Every evaluation in this paper still uses the full, harder settings from Table~\ref{tab:attack-config}.

\subsubsection{\textup{\textbf{Vector-Quantized Codebook}}}
\label{sec:codebook}

We were also motivated by prior work showing that vector quantization can improve robustness. Mao et al.~\cite{mao2021discrete} showed that adding discrete, VQ-based representations to a Vision Transformer makes it more robust to corruptions and adversarial attacks. Yang et al.~\cite{yang2023vector} used a VQ codebook with self-attention to learn features that stay useful across different corruption types, though it does so by training jointly on clean and corrupted images rather than relying on the codebook alone, and He and Singhal~\cite{he2024vqunet} showed that a VQ-based U-Net can filter adversarial noise from images before classification, using only clean training data. These results suggest that snapping an input to one of a small set of learned codebook vectors can remove small perturbations by construction, raising the question of whether this holds without ever needing to see a perturbed example during training. We test whether this same idea, applied to our predictor instead of to raw images or a large trainable backbone, is enough to produce robustness on its own.

This method changes the predictor's architecture instead of its training data. Rather than pooling the encoder's output into a single vector, we work directly on its un-pooled patch tokens $Z \in \mathbb{R}^{N \times D}$, where $N$ is the number of spatial patches ($N{=}196$ for CheXWorld and X-WIN, $N{=}256$ for RadJEPA and I-JEPA) and $D$ is the encoder's embedding width. We reshape $Z$ into a spatial grid $X \in \mathbb{R}^{D \times g \times g}$, where $g = \sqrt{N}$, and pass it through one strided convolution:
\begin{equation}
H = \sigma\big(\text{BN}(\text{Conv}_{3\times3,\,s=2}(X))\big) \in \mathbb{R}^{128 \times g' \times g'}
\end{equation}
This $3{\times}3$, stride-2 convolution maps the $D$ input channels down to 128 and halves the spatial size: a $14{\times}14$ grid ($N{=}196$) becomes $7{\times}7$, and a $16{\times}16$ grid ($N{=}256$) becomes $8{\times}8$. BN is batch normalization and $\sigma$ is GELU.

Each of the $g' \times g'$ spatial positions is quantized separately. For position $i$, its 128-dimensional feature $h_i$ is normalized to $z_{e,i} = \text{LayerNorm}(h_i)$, then matched to the nearest of the codebook's 128 learned vectors:
\begin{equation}
q_i = e_{k_i}, \quad k_i = \arg\min_j \lVert z_{e,i} - e_j \rVert
\end{equation}
This gives 49 lookups per image for the $14{\times}14$-grid encoders and 64 for the $16{\times}16$-grid encoders.

\begin{figure}[t]
  \centering
  \includegraphics[width=1\linewidth]{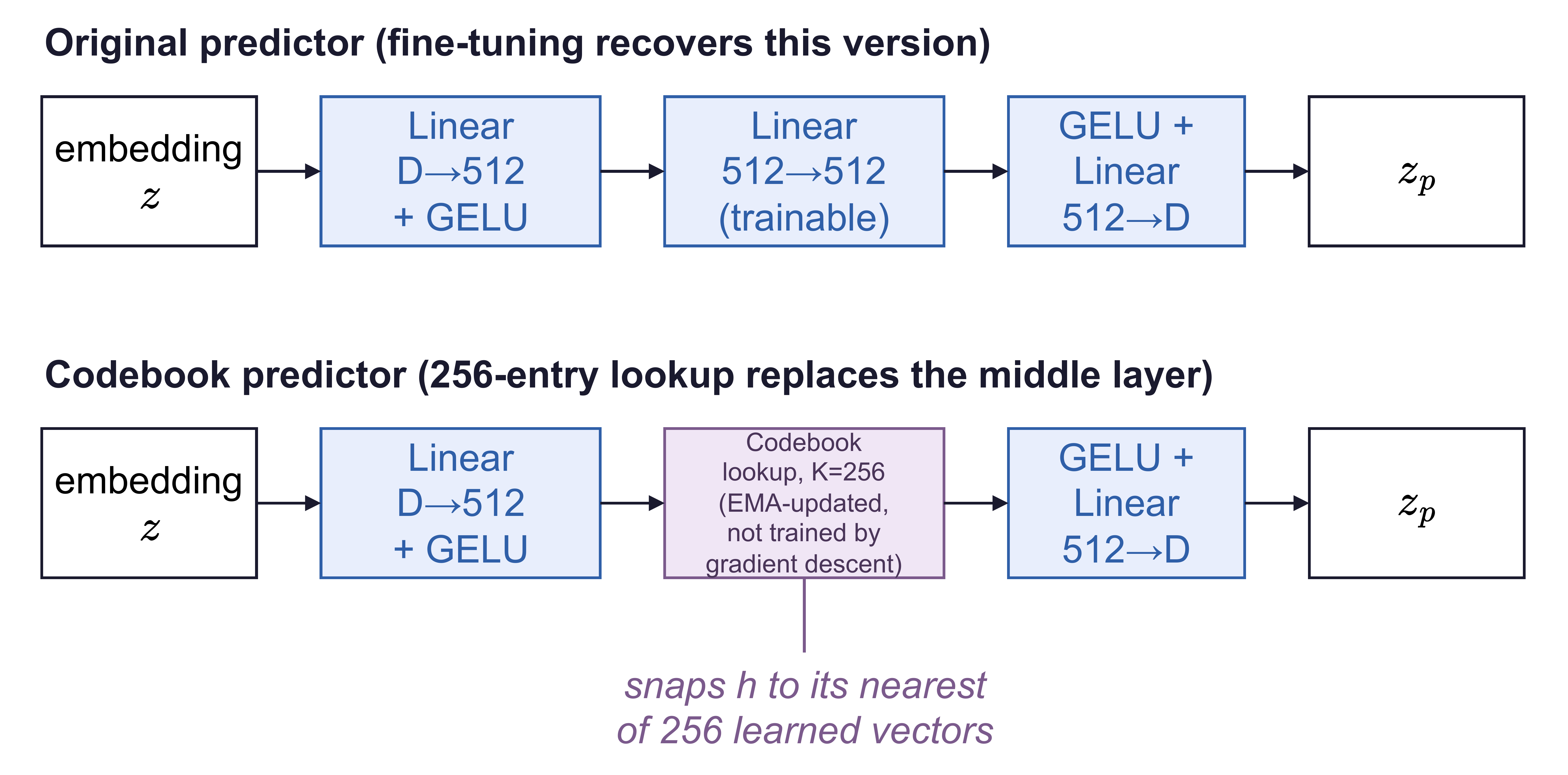}
  \caption{The codebook predictor: tokens are reshaped into a spatial grid, convolved, quantized at each position, and combined with the pre-quantization features via a residual connection before the classifier.}
  \Description{Diagram of the codebook predictor pipeline: token grid reshaping, convolution, per-position vector quantization with a codebook, residual addition, flattening, and a final classifier.}

  \label{fig:codebook}
\end{figure}

Picking the nearest codebook vector is a hard, discrete choice, so it has no gradient. We use a straight-through estimator, $q_i^{\text{ST}} = z_{e,i} + \text{sg}[q_i - z_{e,i}]$, so the forward pass uses the real codebook vector while the backward pass lets gradients skip over this step untouched. Figure~\ref{fig:codebook} shows the full predictor with this step included.

The codebook starts from $k$-means clusters, not random values: we run the convolution and normalisation step on a 1,000-image subset of the clean training set and cluster the resulting vectors into 128 groups, whose centers become the codebook. During training, each codebook entry is nudged toward whatever vectors get assigned to it, using an exponential moving average (decay 0.99), and entries that are rarely used get periodically reset so the whole codebook stays useful.

The quantized feature map is then added back to the pre-quantization convolution output and flattened straight into one linear classifier layer:
\begin{equation}
F = H + Q, \qquad \hat{y} = W_c\, \text{flatten}(F) + b_c
\end{equation}
$Q$ collects all the $q_i^{\text{ST}}$ vectors back into their grid positions. This single linear layer maps the flattened features (6272 or 8192 numbers, depending on the encoder) directly to the 14 disease outputs, with no hidden layer.

This predictor never sees a perturbed image during training, so its loss is just a classification term plus a commitment term that keeps the pre-quantized features close to their assigned codebook entries:
\begin{equation}
\label{eq:vqloss}
\mathcal{L}_{\text{codebook}} = \text{BCE}(y, \hat{y}) + \frac{\beta}{128 \cdot g'^2}\sum_{i=1}^{g'^2} \lVert z_{e,i} - \text{sg}[q_i] \rVert^2
\end{equation}
Here $\beta = 0.25$ is the commitment weight, and $\text{sg}[\cdot]$ blocks gradients so this term only shapes the encoder side, not the codebook itself; the codebook is updated separately, by the EMA step described above.

We train this predictor for 300 epochs with Adam (learning rate $10^{-3}$, weight decay $10^{-4}$, cosine-annealed to $10^{-5}$, batch size 128) on the full 78,468-image clean training set, then evaluate it for the first time on the NIH ChestX-ray14 test split (3,000 images), using the same perturbation tests as fine-tuning (Table~\ref{tab:attack-config}). This tests one question: can the codebook's discrete structure make the model resistant to perturbations through architecture alone, without ever training on a perturbed example? Results are given in Section~5.2, alongside fine-tuning's results in Section~5.3.

\section{Experimental Setup}
\label{sec:setup}

\textbf{Dataset.} We use the official NIH ChestX-ray14 dataset~\cite{wang2017chestx}, using its official train/test split throughout: 78{,}468 images for training, and a separate held-out test set for evaluation. Testing uses distorted versions of the held-out test images created by applying the perturbation tests in Table~\ref{tab:attack-config} to them so that evaluation reflects how the models behave on damaged, real-world-like inputs rather than clean ones.

\textbf{Models.} We test four frozen JEPA-style encoders. Three of them, CheXWorld~\cite{yue2025chexworld}, RadJEPA~\cite{khan2026radjepa}, and I-JEPA~\cite{assran2023self}, are used exactly as publicly released. X-WIN~\cite{yang2026x} is the exception: its original pretrained checkpoint is not publicly available, and its original training recipe uses the NLST (National Lung Screening Trial) dataset \cite{national2011national}, which is too large for us to pretrain on given our compute constraints. We therefore pretrained our own X-WIN encoder on the smaller LIDC-IDRI CT dataset \cite{armato2011lung} instead, for 100 epochs. Once pretrained, this encoder is frozen exactly like the other three and used for all further testing. I-JEPA is the only encoder pretrained on general natural images (ImageNet) rather than medical images, which lets us ask whether a strong general-purpose encoder transfers well to chest X-rays. None of the four encoders' weights are ever changed after pretraining; only the small predictor and classifier on top of them are trained.


\begin{figure*}[t]
\centering
\begin{subfigure}{0.49\textwidth}
  \includegraphics[width=\linewidth]{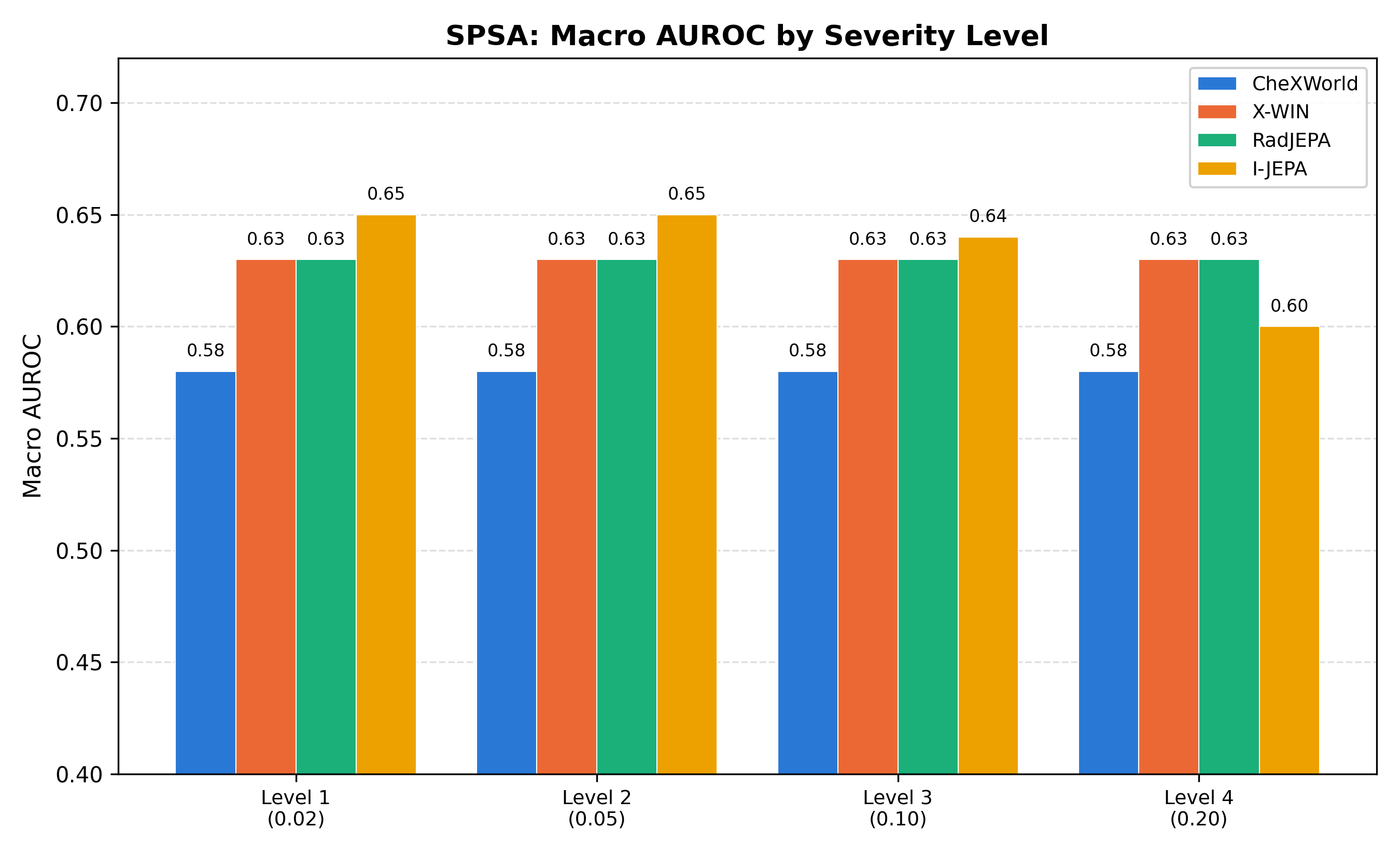}
  \caption{SPSA}
\end{subfigure}
\begin{subfigure}{0.49\textwidth}
  \includegraphics[width=\linewidth]{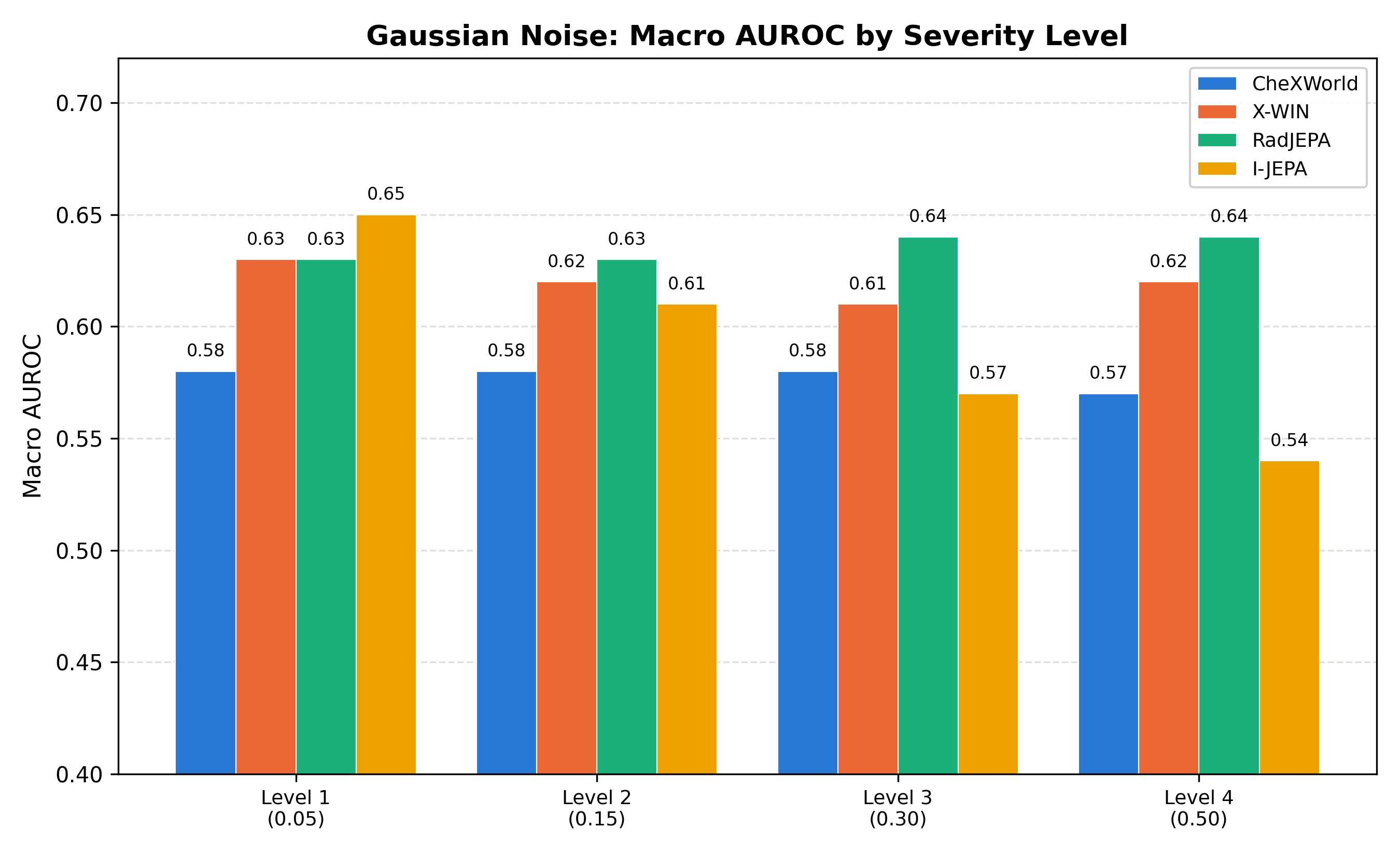}
  \caption{Gaussian Noise}
\end{subfigure}
\begin{subfigure}{0.49\textwidth}
  \includegraphics[width=\linewidth]{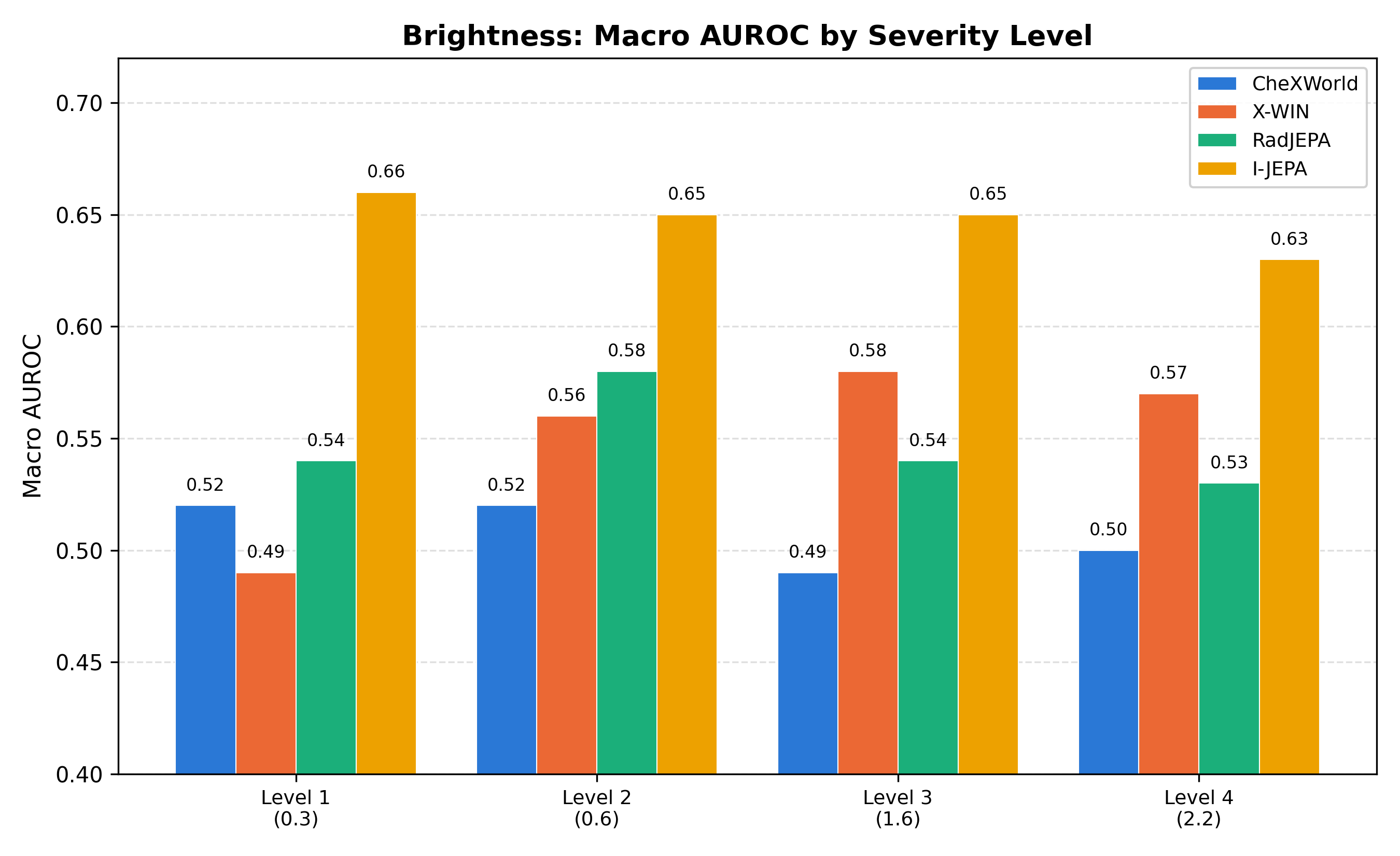}
  \caption{Brightness}
\end{subfigure}
\begin{subfigure}{0.49\textwidth}
  \includegraphics[width=\linewidth]{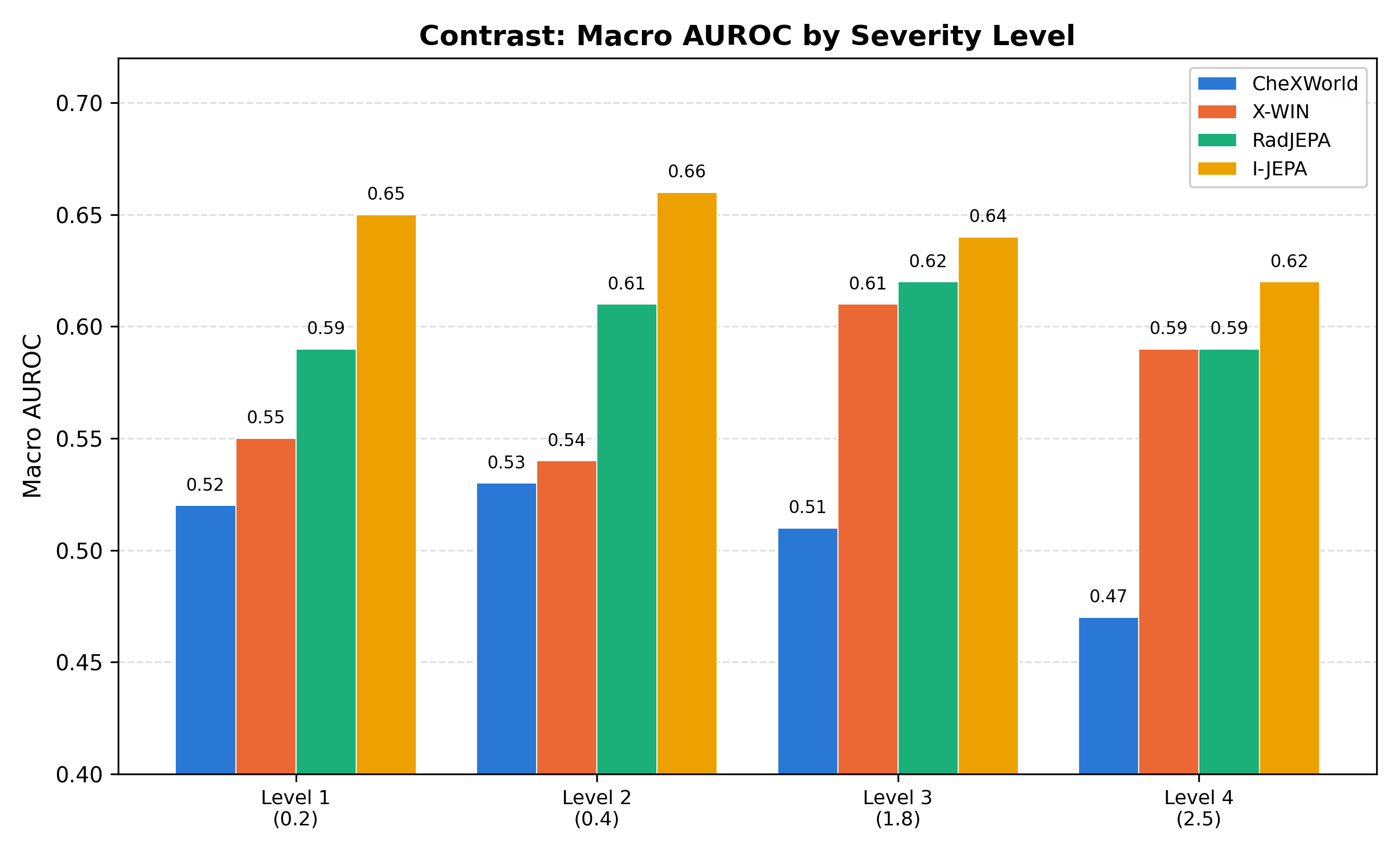}
  \caption{Contrast}
\end{subfigure}
\caption{Macro AUROC across all four severity levels of each perturbation test, each model, on the 1,000-image held-out test subset. }
\Description{Four grouped bar charts, one per perturbation test (SPSA, Gaussian Noise, Brightness, Contrast), each showing Macro AUROC across four severity levels for CheXWorld, X-WIN, RadJEPA, and I-JEPA.}
\label{fig:level-degradation-auroc}
\end{figure*}

\textbf{Training.} The predictor and classifier use the Adam optimiser (weight decay $10^{-4}$). The clean predictor and classifier are trained for 300 epochs at learning rate $10^{-2}$ on all 78{,}468 clean training images; this pair is the ``Original'' baseline. For adversarial fine-tuning, the predictor and classifier are warm-started from these clean-trained weights rather than trained from scratch, then trained for 200 epochs at learning rate $10^{-3}$ on the full training split augmented with its corrupted versions, plus a large subset used for the more expensive SPSA/Boundary attacks, so that classification and embedding consistency improve together (Equation~\ref{eq:ftloss}). The codebook variant instead uses a separate convolutional classifier with a 128-entry codebook (Section~\ref{sec:codebook}), trained with the Adam optimiser (learning rate $10^{-3}$, weight decay $10^{-4}$, cosine-annealed to $10^{-5}$) for 300 epochs in mini-batches of 128, on the same 78,468 clean training images only; no perturbed images are used during codebook training (Equation~\ref{eq:vqloss}).

\textbf{Evaluation metrics.} Every result in Section~\ref{sec:results} uses two metrics: Macro AUROC (the per-class area under the ROC curve, averaged over the 14 labels) and Embedding Cosine Similarity between each image's clean and perturbed predictor output (Equation~\ref{eq:cossim}). These are evaluation-only metrics; they are never optimised directly. The training objectives are Equation~\ref{eq:ftloss} for the clean and fine-tuned predictors and Equation~\ref{eq:vqloss} for the codebook predictor.

\textbf{Hardware.} All experiments run on two NVIDIA RTX A6000 GPUs (48\,GB each) in parallel, with two encoders assigned to each GPU.

\section{Results}
\label{sec:results}

\subsection{How Much Damage Does Each Test Cause?}
\label{sec:before}
\begin{figure*}[t]
\centering
\begin{subfigure}{0.49\textwidth}
  \includegraphics[width=\linewidth]{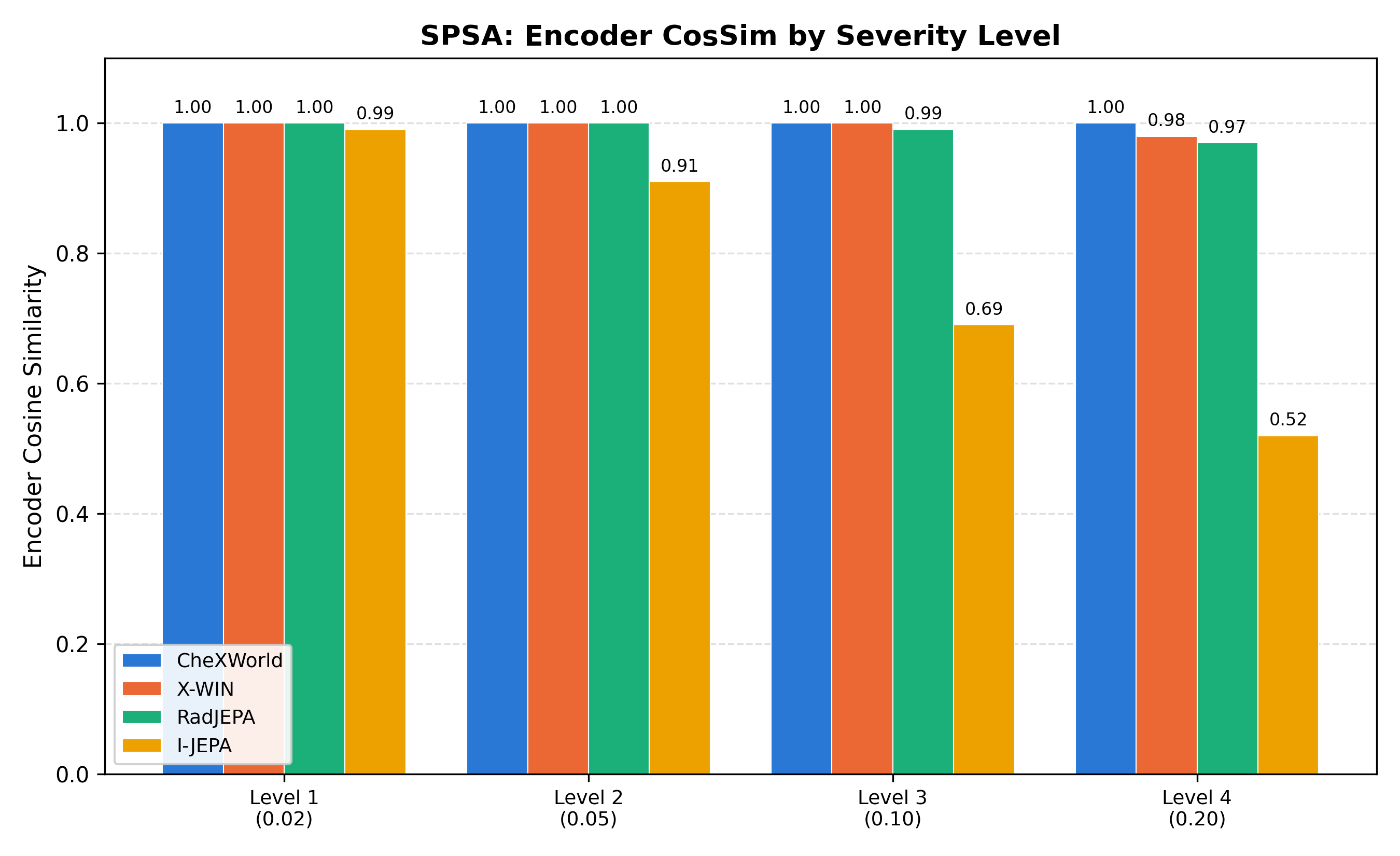}
  \caption{SPSA}
\end{subfigure}
\begin{subfigure}{0.49\textwidth}
  \includegraphics[width=\linewidth]{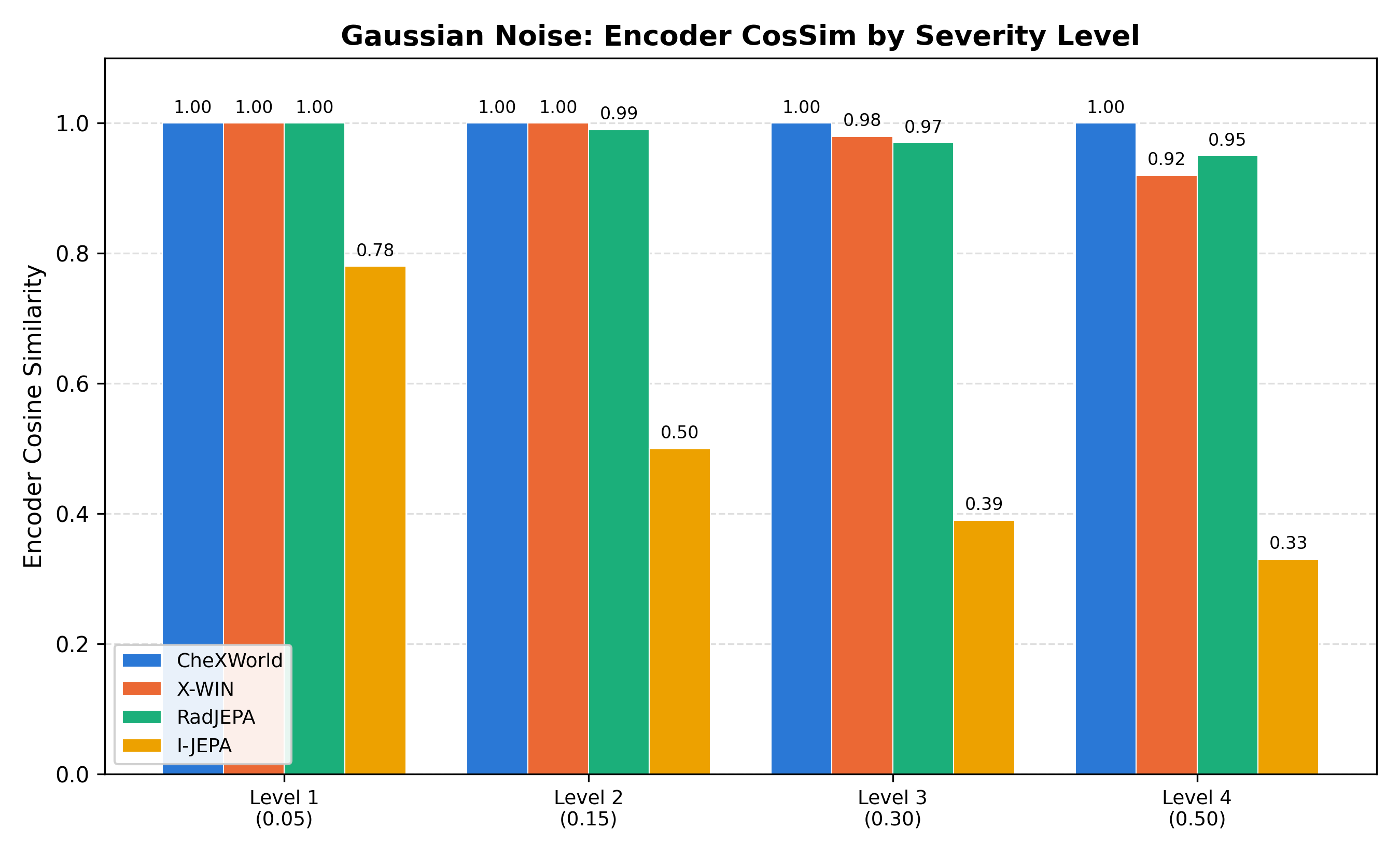}
  \caption{Gaussian Noise}
\end{subfigure}
\begin{subfigure}{0.49\textwidth}
  \includegraphics[width=\linewidth]{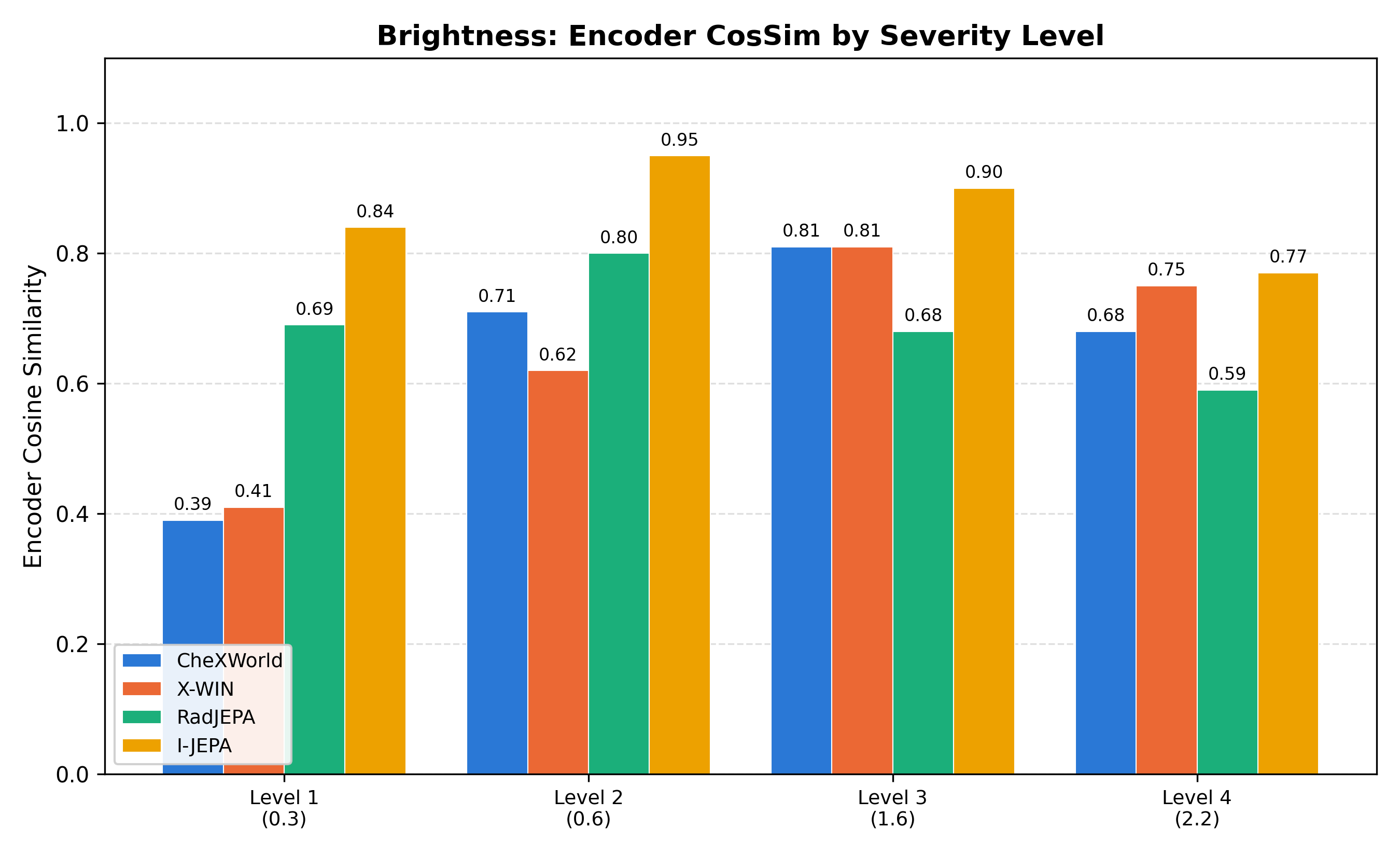}
  \caption{Brightness}
\end{subfigure}
\begin{subfigure}{0.49\textwidth}
  \includegraphics[width=\linewidth]{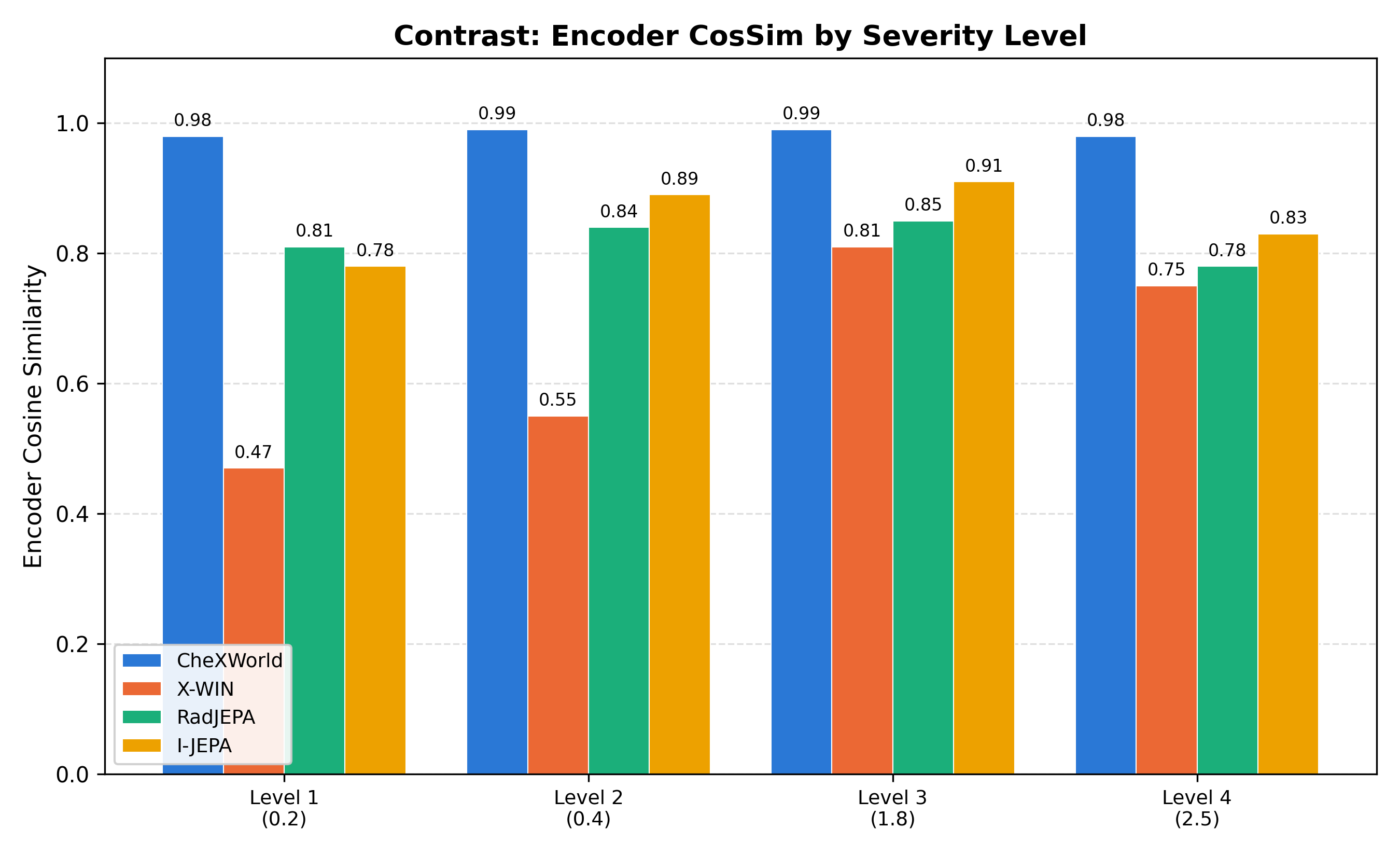}
  \caption{Contrast}
\end{subfigure}
\caption{Encoder-level Embedding Cosine Similarity across all four severity levels of each perturbation test, each model, on the same held-out test subset.}
\Description{Four grouped bar charts, one per perturbation test, each showing encoder-level cosine similarity across four severity levels for CheXWorld, X-WIN, RadJEPA, and I-JEPA.}
\label{fig:level-degradation-cossim}
\end{figure*}

Figures~\ref{fig:level-degradation-auroc} and~\ref{fig:level-degradation-cossim} show how Macro AUROC and encoder-level CosSim move as each perturbation gets stronger, across all four severity levels. Table~\ref{tab:level-degradation-pct} summarises this at the strongest level for the predictor without a codebook, on average, CheXWorld degrades by 6.90\%, X-WIN by 3.17\%, RadJEPA by 4.76\%, and I-JEPA by 10.77\%. Under SPSA and Gaussian Noise, CheXWorld, X-WIN, and RadJEPA stay almost flat, while I-JEPA degrades steadily, its Gaussian-Noise CosSim collapsing to 0.33. Brightness and Contrast do not get steadily worse, embedding similarity is often lowest at the \emph{mildest} setting (CheXWorld and X-WIN's CosSim drops to 0.39 and 0.41 at Level~1), even though AUROC keeps falling through Level~4. The Boundary Test behaves like SPSA and Noise, with I-JEPA showing its sharpest single-test drop (18.46\%, CosSim to 0.39).
For RadJEPA under Gaussian Noise, the table shows $-1.59\%$, meaning AUROC did not drop at all, it was actually a tiny bit higher on the noisy images (0.64) than on the clean ones (0.63). This does not mean noise helped the model; with only 1,000 test images, a difference this small can happen just by chance. Each test damages a different model the most, so no single test tells the whole story.

\begin{table}[t]
\centering
\caption{Macro AUROC at the strongest severity level of each perturbation test, each model, on the 1,000-image held-out test subset.} 
\label{tab:level-degradation-pct}
\resizebox{\columnwidth}{!}{%
\footnotesize
\begin{tabular}{l cccc}
\toprule
Test & CheXWorld & X-WIN & RadJEPA & I-JEPA \\
\midrule
\textbf{Clean} & 0.58 & 0.63 & 0.63 & 0.65 \\
\midrule
SPSA (0.20) & 0.58 (0.00\%) & 0.63 (0.00\%) & 0.63 (0.00\%) & 0.60 (7.69\%) \\
Boundary (80 q) & 0.57 (1.72\%) & 0.63 (0.00\%) & 0.63 (0.00\%) & 0.53 (18.46\%) \\
Gaussian Noise (0.50) & 0.57 (1.72\%) & 0.62 (1.59\%) & 0.64 ($-1.59\%$) & 0.54 (16.92\%) \\
Brightness (2.2) & 0.50 (13.79\%) & 0.57 (9.52\%) & 0.53 (15.87\%) & 0.63 (3.08\%) \\
Contrast (2.5) & 0.47 (18.97\%) & 0.59 (6.35\%) & 0.59 (6.35\%) & 0.62 (4.62\%) \\
\midrule
\textbf{Average} & \textbf{0.54 (6.90\%)} & \textbf{0.61 (3.17\%)} & \textbf{0.60 (4.76\%)} & \textbf{0.58 (10.77\%)} \\

\bottomrule
\end{tabular}%
}
\end{table}

\subsection{Codebook Test: Results}
\label{sec:codebook-results}
Table~\ref{tab:codebook-degradation-pct} reports Macro AUROC for the 128-entry codebook at the strongest severity level. On average, CheXWorld degrades by 10.00\%, X-WIN by 13.56\%, RadJEPA by 9.38\%, and I-JEPA by 21.43\%. RadJEPA and CheXWorld hold up best; I-JEPA is hit hardest, dropping by as much as 24.29\% under Gaussian Noise.

This falls short of the gains reported for vector quantization in prior work~\cite{yang2023vector,he2024vqunet}, likely because our setting is more constrained: our codebook is trained on clean images only, sits inside a small predictor rather than a large trainable backbone, has only 128 entries, and operates on top of an encoder that is never updated. With so little capacity to work with, a discrete bottleneck alone is not enough to resist perturbations it was never trained on.

\begin{table}[t]
\centering
\caption{Codebook Macro AUROC at the strongest severity level of each perturbation test, each model, on the 3,000-image held-out test subset.}
\label{tab:codebook-degradation-pct}
\resizebox{\columnwidth}{!}{%
\footnotesize
\begin{tabular}{l cccc}
\toprule
Test & CheXWorld & X-WIN & RadJEPA & I-JEPA \\
\midrule
\textbf{Clean} & 0.70 & 0.59 & 0.64 & 0.70 \\
\midrule
SPSA (0.20) & 0.63 (10.00\%) & 0.50 (15.25\%) & 0.58 (9.38\%) & 0.56 (20.00\%) \\
Boundary (80 q) & 0.65 (7.14\%) & 0.51 (13.56\%) & 0.59 (7.81\%) & 0.57 (18.57\%) \\
Gaussian Noise (0.50) & 0.61 (12.86\%) & 0.50 (15.25\%) & 0.57 (10.94\%) & 0.53 (24.29\%) \\
Brightness (2.2) & 0.62 (11.43\%) & 0.51 (13.56\%) & 0.58 (9.38\%) & 0.55 (21.43\%) \\
Contrast (2.5) & 0.63 (10.00\%) & 0.51 (13.56\%) & 0.58 (9.38\%) & 0.54 (22.86\%) \\
\midrule
\textbf{Average} & \textbf{0.63 (10.00\%)} & \textbf{0.51 (13.56\%)} & \textbf{0.58 (9.38\%)} & \textbf{0.55 (21.43\%)} \\

\bottomrule
\end{tabular}%
}
\end{table}

\subsection{Fine-Tuning: Results}
\label{sec:ft-results}
Table~\ref{tab:finetune-auroc-worst} shows fine-tuned Macro AUROC at the strongest level of each test. On average across the five tests, fine-tuning reaches 0.62 for CheXWorld, 0.63 for X-WIN, 0.62 for RadJEPA, and 0.68 for I-JEPA, gains that hold across three of the four models, with I-JEPA (the most damaged model, per Section~\ref{sec:before}) improving the most and RadJEPA (the least damaged) essentially unchanged, consistent with its already-high robustness.

\begin{table}[t]
\centering
\caption{Fine-Tuned Macro AUROC at the strongest severity level of each perturbation test, each model, on the held-out test subset.}
\label{tab:finetune-auroc-worst}
\resizebox{\columnwidth}{!}{%
\footnotesize
\begin{tabular}{l cccc}
\toprule
Test & CheXWorld & X-WIN & RadJEPA & I-JEPA \\
\midrule
SPSA (0.20)         & 0.65 & 0.66 & 0.65 & 0.68 \\
Boundary (80 q)     & 0.65 & 0.65 & 0.65 & 0.70 \\
Gaussian Noise (0.50) & 0.64 & 0.65 & 0.65 & 0.68 \\
Brightness (2.2)    & 0.55 & 0.59 & 0.53 & 0.66 \\
Contrast (2.5)      & 0.60 & 0.63 & 0.60 & 0.67 \\
\midrule
\textbf{Average}    & \textbf{0.62} & \textbf{0.63} & \textbf{0.62} & \textbf{0.68} \\

\bottomrule
\end{tabular}%
}
\end{table}



\subsection{Few Sample Robustness: How Label-Efficient Are the Raw Embeddings?}
\label{sec:lowlabel}

\begin{figure}[t]
  \centering
  \includegraphics[width=0.75\linewidth]{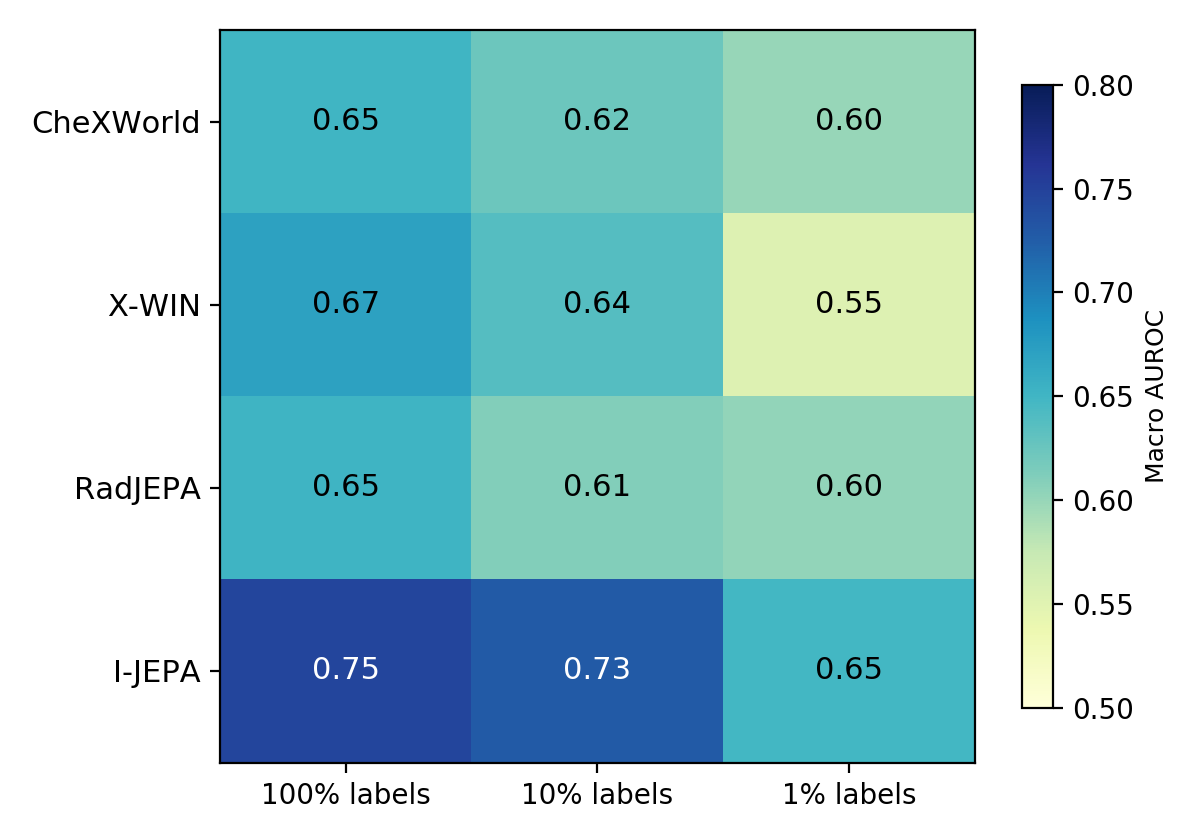}
  \caption{Few-sample robustness of the raw (Original-predictor) embeddings: Macro AUROC on the held-out test set when a fresh linear classifier is trained on 100\%, 10\%, or 1\% of the training labels. Darker means higher AUROC.}
  \Description{Heatmap with four encoder rows (CheXWorld, X-WIN, RadJEPA, I-JEPA) and three columns (100\%, 10\%, 1\% labels) showing Macro AUROC values; I-JEPA is highest across all columns.}
  \label{fig:lowlabel}
\end{figure}

The tests above measure robustness to input perturbations. Here we ask a simpler question, how good are the encoders' raw embeddings when very little labeled data is available? For each model we train a fresh linear classifier on clean images only, using 100\%, 10\%, or 1\% of the training labels, and evaluate on the held-out set. Figure~\ref{fig:lowlabel} reports the Macro AUROC of these Original-predictor embeddings.

Two things stand out. First, I-JEPA has the strongest embeddings at every budget (0.75, 0.73, 0.65), even though it was pretrained on natural images rather than chest X-rays, so a strong general-purpose encoder transfers surprisingly well. Second, the models differ in how gracefully they degrade as labels shrink. X-WIN starts highest among the medical encoders at full labels (0.67) but falls the most at 1\% (0.55), while CheXWorld and RadJEPA are lower at full labels yet hold up better at 1\% (both 0.60). We also passed clean images through the fine-tuned and codebook predictors and found that this training does not reliably improve low-label AUROC; it changes what an embedding is good at rather than uniformly helping. Training aimed at perturbation robustness should therefore not be assumed to also help label-efficient learning.

\subsection{Fine-Tuning Across Random Seeds}
\label{sec:multiseed}

\begin{figure}[t]
  \centering
  \includegraphics[width=0.75\linewidth]{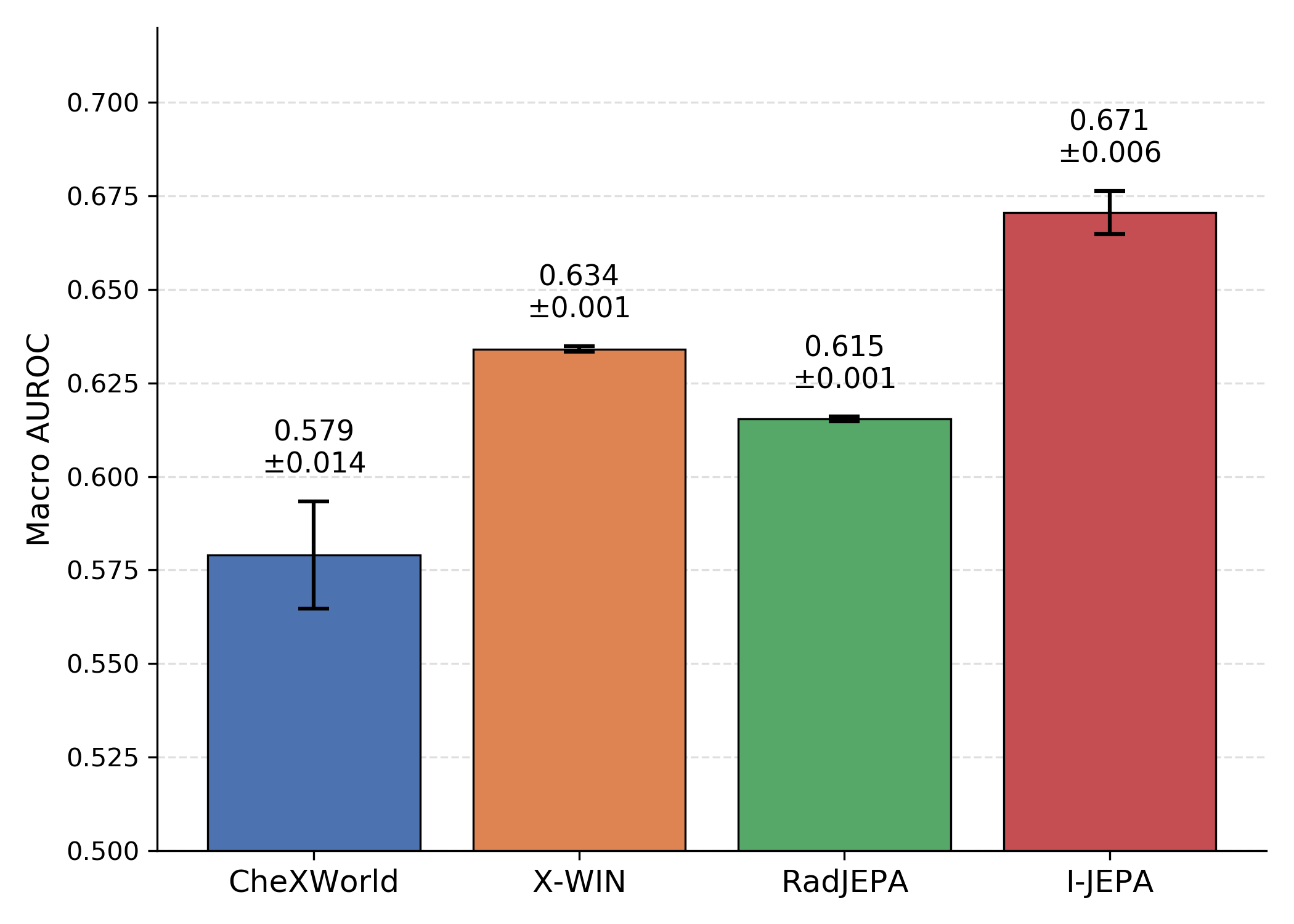}
  \caption{Average fine-tuned Macro AUROC across all five tests (strongest level), per model, under three random seeds.}
  \Description{Grouped bar chart with four model groups (CheXWorld, X-WIN, RadJEPA, I-JEPA), each with three bars for seeds 42, 123, and 456, showing average fine-tuned Macro AUROC. X-WIN, RadJEPA, and I-JEPA are consistent across seeds; CheXWorld varies more.}
  \label{fig:seed-comparison}
\end{figure}

The fine-tuning results in Section~\ref{sec:ft-results} use a single random seed (42). To check whether they depend on that initialisation, we repeat fine-tuning under three more seeds (123, 456, 789), keeping every other setting identical, including the 3,000-image test subset (Table~\ref{tab:attack-config}, Section~\ref{sec:ft}), and compare against the seed-42 baseline on the same subset. Figure~\ref{fig:seed-comparison} reports the mean and standard deviation of fine-tuned Macro AUROC across all five tests, per model, over the three repeat seeds.

X-WIN and RadJEPA are essentially seed-invariant (std $\leq 0.001$), matching their seed-42 baselines closely (0.633 and 0.616). I-JEPA varies slightly more (std $= 0.006$, mean 0.671 vs.\ seed-42's 0.677). CheXWorld is the most seed-sensitive (std $= 0.014$), with its repeat-seed mean (0.579) noticeably below its seed-42 result (0.616). Overall, this supports that the fine-tuning gains in Section~\ref{sec:ft-results} are not an artifact of one initialisation, though CheXWorld's result should be read with its higher seed-sensitivity in mind.

\section{Conclusion}

Across four JEPA-based chest X-ray encoders and six robustness tests, we found sharp, model-dependent degradation, with embedding similarity falling to 0.33 in the worst case (Figure~\ref{fig:level-degradation-cossim}), none of these encoders is robust to realistic corruptions or black-box attacks by default. Fine-tuning the predictor on clean and perturbed images together improved AUROC for most models and tests (Table~\ref{tab:finetune-auroc-worst}), and this held across three random seeds (Figure~\ref{fig:seed-comparison}). The Vector-Quantised codebook, trained on clean data only, told a different story: it fell below its own clean-image performance for every encoder, and degraded more than the plain Original predictor (Table~\ref{tab:codebook-degradation-pct}),  a discrete bottleneck alone was not enough without training on perturbed examples. Robustness should not be assumed from pretraining or architecture; it has to be measured directly.

\bibliographystyle{ACM-Reference-Format}
\bibliography{paper-refs}

\end{document}